\documentclass[letterpaper, 10 pt, conference]{ieeeconf}  % Comment this line out if you need a4paper

\IEEEoverridecommandlockouts                              % This command is only needed if 
\usepackage{graphics} % for pdf, bitmapped graphics files
\usepackage{epsfig} % for postscript graphics files
\usepackage{amsmath} % assumes amsmath package installed
\usepackage{amssymb}  % assumes amsmath package installed
\let\labelindent\relax
\usepackage{enumitem}
\usepackage{todonotes}
\usepackage{url}
\usepackage[scriptsize]{subfigure}
\usepackage[nolist]{acronym}

\DeclareMathOperator{\sign}{sgn}

\newacro{sj}[SJ]{Shared Control-Joystick}
\newacro{sw}[SW]{Switching}
\newacro{sg}[SG]{Shared Control-Gesture}
\newacro{vr}[VR]{Virtual Reality}
\newacro{hmd}[HMD]{Head-Mounted Display}
\newacro{dwa}[DWA]{Dynamic Window Approach}
\newacro{ros2}[ROS2]{Robot Operating System 2}
\newacro{hidwa}[HI-DWA]{Human-Influenced Dynamic Window Approach}

\newcommand{\reals}{\mathbb{R}}
\newcommand{\robot}{\mathcal{A}}

\title{\LARGE \bf
HI-DWA: Human-Influenced Dynamic Window Approach for\\ Shared Control of a Telepresence Robot
}

\author{Juho Kalliokoski, Basak Sakcak, Markku Suomalainen, Katherine J. Mimnaugh,\\ Alexis P. Chambers, Timo Ojala, and Steven M. LaValle % <-this % stops a space
\thanks{*This work was supported by a European Research Council Advanced Grant (ERC AdG, ILLUSIVE: Foundations of Perception Engineering, 101020977), Academy of Finland (projects PERCEPT 322637, CHiMP 342556), and Business Finland (project HUMOR 3656/31/2019).}% <-this % stops a space
\thanks{Authors are with Center of Ubiquitous
Computing, Faculty of Information Technology and Electrical Engineering,
University of Oulu, Finland. {\tt\small \{name.surname\}@oulu.fi}}%
}

\begin{document}

\maketitle
\thispagestyle{empty}
\pagestyle{empty}

%%%%%%%%%%%%%%%%%%%%%%%%%%%%%%%%%%%%%%%%%%%%%%%%%%%%%%%%%%%%%%%%%%%%%%%%%%%%%%%%
\begin{abstract}
This paper considers the problem of enabling the user to modify the path of a telepresence robot. The robot is capable of autonomously navigating to a goal predefined by the user, but the user might still want to modify the path, for example, to go further away from other people, or to go closer to landmarks she wants to see on the way. We propose \ac{hidwa}, a shared control method aimed for telepresence robots based on \ac{dwa} that allows the user to influence the control input given to the robot.
To verify the proposed method, we performed a user study (N=32) in \ac{vr} to compare \ac{hidwa} with switching between autonomous navigation and manual control for controlling a simulated telepresence robot moving in a virtual environment. Results showed that users reached their goal faster using \ac{hidwa} controller and found it easier to use. Preference between  the two methods was split equally. Qualitative analysis revealed that a major reason for the participants that preferred switching between two modes was the feeling of control. We also analyzed the effect of different input methods, joystick, and gesture, on the preference and perceived workload. 

%This paper proposes \ac{hidwa}, a shared control method aimed for telepresence robots enabling the user to influence the control input chosen by \ac{dwa}.
%where the user can influence the trajectory chosen by a \ac{dwa}. 
%The underlying idea is that when a telepresence robot has semi-autonomous navigation, such that the user can choose the robot's goal position, she might still want to modify the trajectory without changing the goal, for example, to go further away from other people than the autonomous navigation would, or go closer to signposts she wants to read on the way. 
%To verify the proposed method, we performed a user study (N=32) in \ac{vr} where a simulated telepresence robot autonomously traversed a virtual environment with the user aboard the robot. We implemented the proposed method with two input methods from the user, a joystick and gesture control, and used a possibility to switch between fully manual joysticking and fully autonomous motion as a comparison. Results showed that users reached their goal faster with the proposed method and found it easier, even though preference between the two methods was split equally. Qualitative analysis revealed a major reason for many users preferring the comparison method was the feeling of control.%, which needs further refinements in future studies. 
\end{abstract}

\section{Introduction}
\label{sec:intro}
Telepresence robots are one of the most prominent tools that can enable \textit{hybrid meetings} such that part of the people are physically present but others who wish to participate can only do so virtually. The use cases range from important personal milestones (birthdays, weddings, graduations) to business meetings and factory tours. The commercial%, ``regular" 
telepresence robots, essentially Skype-on-wheels, are sufficient for video conferencing. However, a recent study showed that in a hybrid meeting, virtually present people did not speak as much and found a shared task more difficult compared to physically present people \cite{stoll2018wait}. These limitations increased the interest in immersive robotic telepresence for which the user embodies a mobile robot in a remote location through a \ac{hmd}.
Despite being more complex and prone to issues such as VR sickness \cite{Lav00,LaValle_bookVR}, the immersiveness of an \ac{hmd} has the potential to overcome the gap between physically and virtually present people.

%however, lately interest has increased on making the remote user wear an \ac{hmd}. Despite being more complex and prone to issues such as VR sickness \cite{Lav00,LaValle_bookVR}, the immersiveness of an \ac{hmd} has the potential to overcome the gap between physically and virtually present people, reported in \cite{stoll2018wait}, where users virtually present people did not speak as much and found a shared task more difficult than physically present people. 
%telepresence robots, essentially Skype-on-wheels, which are as easy to use for video conferencing; however, lately interest has increased on making the remote user wear an \ac{hmd}. Despite being more complex and prone to issues such as VR sickness \cite{Lav00,LaValle_bookVR}, the immersiveness of an \ac{hmd} has the potential to overcome the gap between physically and virtually present people, reported in \cite{stoll2018wait}, where users virtually present people did not speak as much and found a shared task more difficult than physically present people. 

Even though shared control is a well-known problem in robotics, there is limited research on aspects related to telepresence, either for ``regular" or immersive telepresence. Manually controlling the robot through joystick commands is the simplest method to enable the user to affect the robot motion. However, there is evidence that people find it tiring \cite{rae2017robotic} leading to the research towards autonomous or semi-autonomous methods, such as the ``waypoint navigation" found in the commercial telepresence robot Double for which the user points at a point within the visible area and the robot navigates towards there autonomously. This type of navigation is also shown to be useful for immersive telepresence \cite{baker2020towards}. However, as telepresence robots share an environment with people, a difficult problem of human-aware navigation \cite{kruse2013human} should be addressed. In addition to safely avoiding people in the environment, a telepresence robot should also ensure that the person on board feels comfortable with the robot motion. Since aspects related to comfort and preference vary from one person to another, it seems natural to endow the user with effortless methods to make slight changes to the robot's autonomous path, to easily address issues such as going too close to people or too far from points of interest not known to the autonomous planner.

%Even though shared control is a well-known problem in robotics, there is a limited amount of research about it for telepresence, either for ``regular" or immersive telepresence.
%Whereas joysticking is the simplest method, there is evidence that people find it tiring \cite{rae2017robotic}, leading the research towards autonomous or semi-autonomous methods, such as the "waypoint control" found in the commercial telepresence robot Double, and also shown useful for immersive telepresence in research \cite{baker2020towards}; here, the user chooses a point within the visible area, where the robot then moves autonomously. 
%However, as telepresence robots often move around people, the autonomous path should address the difficult problem of human-aware navigation \cite{kruse2013human}. 
%Whereas modern human-aware planners work very well, there are also personal preferences, and in addition, also the user aboard the telepresence robot should feel comfortable about the robot's path, not only the people around the robot. Hence, with a user already aboard the robot all the time, it seems natural to endow the user with effortless methods to make slight changes to the robot's autonomous path, to easily address issues such as going too close to people or too far from points of interest not known to the autonomous planner.

\begin{figure}
\centering
\includegraphics[trim=250 275 200 120,clip,width=0.99\columnwidth]{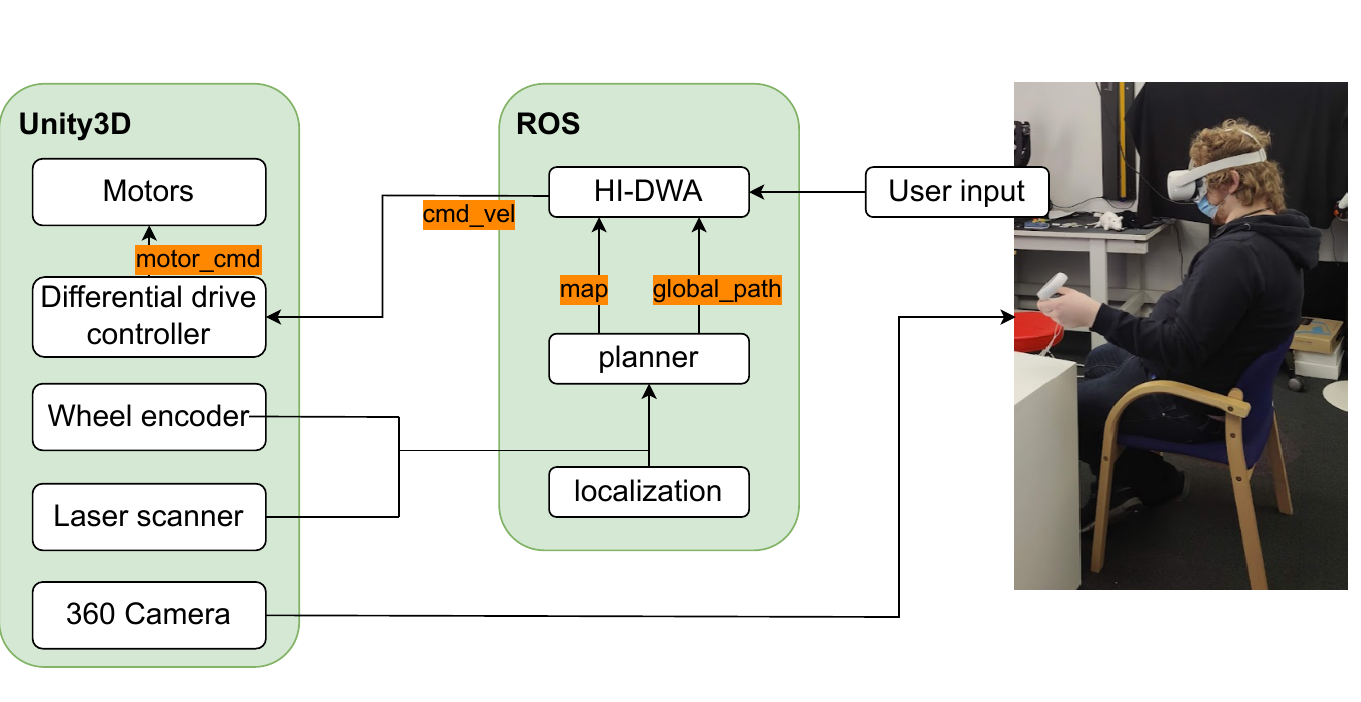}
\caption{Navigation framework for a telepresence robot employing the proposed HI-DWA shared control method and its communication with the simulation environment Unity.}
\label{fig:teleop} 
\vspace{-0.5cm}
\end{figure}

In this paper, we address the problem of semi-autonomous navigation of a telepresence robot that allows the user to influence the trajectory executed by the robot with minimal effort. We propose \ac{hidwa} for robot control, that is an adaptation of the popular \ac{dwa} \cite{fox1997dynamic}, which searches for the best control input that optimizes a relevant objective by integrating the system dynamics for each admissible control input and scoring the resulting trajectories. The key idea of \ac{hidwa} is to penalize the deviation from the user input. Thus, we retain the collision avoidance property of \ac{dwa}, while allowing the user to influence the robot's motion. 
We performed a user study (N=32) in which the participants wearing \ac{hmd}s compared the proposed control method with switching between manual and autonomous modes (a method often used in commercial telepresence robots, such as Double 3) for a simulated robot in a virtual environment that mimics immersive telepresence. They were encouraged to alter the robot's path. The results showed that even though participants found the proposed method easier, there was no difference in preference. Further analyses to interpret this result indicated that a major reason for seeing no difference in preference is that many users liked having full control over the robot's motion.
Also, for the use case of immersive telepresence, we found that adjusting the path using gestures (moving hand) was preferred over joystick.

\section{Related Work}

%%%%https://ieeexplore.ieee.org/document/8931565
%need to talk about dwb/dwa planners?
%https://ieeexplore.ieee.org/document/7759527
%Switching between manual and autonomous resulted in better scores but was perceived harder than completely autonomous movement
%Tele-Operating USAR Robots: Does Driving Performance Increase with Aperture Width or Practice?
%Letting users practice makes them more comfortable with the controls and leads better scores for the task
%https://arxiv.org/abs/2011.05228
%VFH+ based shared control for remotely operated mobile robots. Blended commands coming from VFH+ and the user given trajectory. No mentions about global planning and also only reduced amount of collisions. (Blended trajectory was not checked for collision)

In robotic telepresence a robot in the local environment is connected to and controlled by a user in a remote location~\cite{Kristoffersson2013}. The user is able to move the robot around and interact with other people in the remote environment. The term telepresence was originally coined by Minsky \cite{minskyTele} and the most prominent telepresence implementation in research is a mobile robot with a camera, a standard screen, and some kind of controller~\cite{Kristoffersson2013}. Such telepresence robots are used for, for example, education \cite{botev2021immersive},
social interaction \cite{Kristoffersson2013}, and telemonitoring on the basis of medical consultation \cite{carranza2018akibot}.

%witching the autonomy has been researched a lot \cite{chiou2016experimental} \cite{petousakis2020human}. Another often used method is to blend the trajectory coming from autonomous controller with the trajectory indicated by user \cite{6859352} \cite{8009352} \cite{2020} \cite{8914558} 

Shared control refers to systems in which the human and the robot are cooperating to achieve a specific goal. Whereas shared control is often used with manipulators \cite{zeestraten2018programming} or autonomous cars \cite{marcano2020review}, there is also a growing literature of shared control for autonomous robots. Shared control is often achieved by blending trajectories or policies \cite{dragan2013policy}, using the autonomous controller as a safeguard for detecting collisions, or simply switching between autonomous and manual modes \cite{chiou2016experimental,petousakis2020human}.

In trajectory blending, commands coming from the autonomous controller are blended with the trajectory indicated by the user \cite{6859352} \cite{2020} \cite{8914558} \cite{wang2014adaptive}. However, apart from some of these (\cite{6859352} \cite{wang2014adaptive}), there is no proper collision detection for the resulting trajectory.
In contrast, using the autonomous controller as a safeguard means that normally user has full control of the robot, but if some hazard or obstacle is detected by the robot it will take over the control preventing a collision. This method was already used in lunar rovers \cite{krotkov1996safeguarded} and has been used in many mobile robots afterwards \cite{luo2020teleoperation}\cite{jiang2014shared}\cite{fong2001safeguarded}. The proposed method differs from these by properly addressing the possible obstacles like in a safeguard method, but also taking the advantage of autonomous planning so that the user does not have to control the robot actively.

%In any VR application, VR sickness plays a big role just as in any virtual reality application. A major reason for the VR sickness is vection, which is the illusion of self motion~\cite{LaValle_bookVR}. However the the cause and severeness can vary widely depending on the person and used hardware~\cite{chang2020virtual}. 
%A telepresence robot sets many limitations on avoiding VR sickness. In many VR applications vection is avoided by using teleportation~\cite{buttussi2019locomotion}, but that is not viable  due to the delay it would require when using continuously moving telepresence robot. Having the robot moving autonomously also restricts the users control over the motions, which several studies have shown causing VR sickness~\cite{chang2020virtual}.

%In many VR applications vection is avoided by using teleportation~\cite{buttussi2019locomotion}, but that is not viable  due to the delay it would require when using continuously moving telepresence robot. 

%%%%%%%%%%%%%%%%%%%%%%%%%%%%%%%%%%%%%%%%%%%%%%%%%%%%%%%%%%%
%\section{A human-input aware controller for a mobile telepresence robot}
%\section{A shared control method for a mobile telepresence robot}
\section{Proposed shared control method}
\label{sec:shared_control}

% \begin{figure}
% \centering
% \includegraphics[trim=0 0 0 0,clip,width=0.99\columnwidth]{figures/robot_overview.jpeg}
% \caption{The simulated robot in its environment 
% }
% \label{fig:robot_overview} 
% \vspace{-0.5cm}
% \end{figure}

\begin{figure}
    \centering
    \subfigure[]{\includegraphics[trim=0 0 0 0,clip,width=0.45\linewidth]{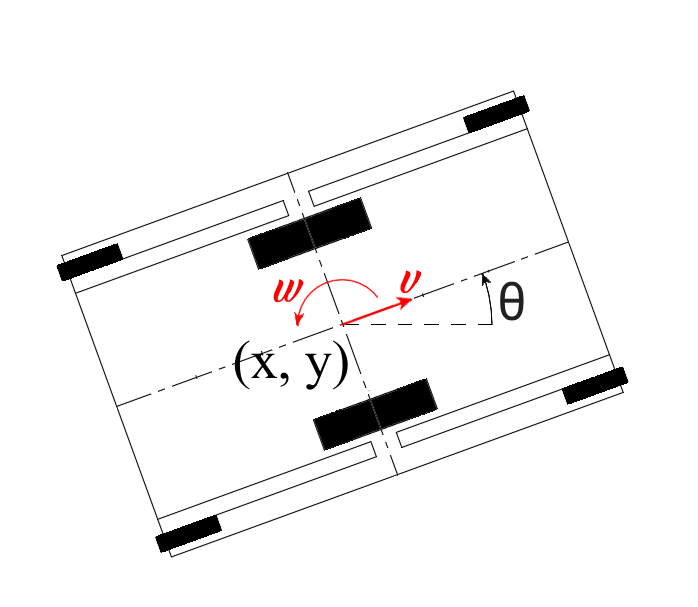}}
    \subfigure[]{\includegraphics[trim=0 0 0 0,clip,width=0.45\linewidth]{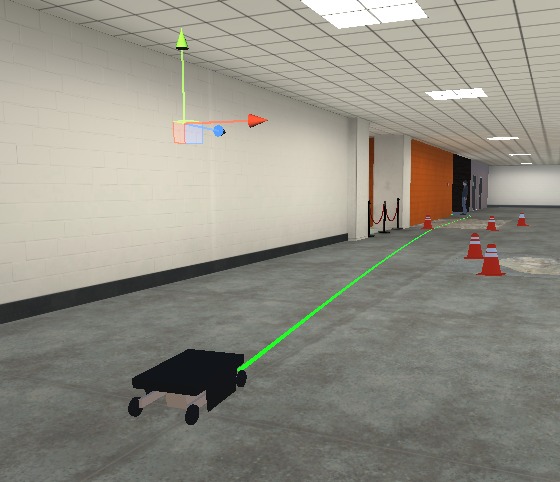}}
    \caption{(a) Diagram of the robot used in the experiment with state and control variables shown. (b) The simulated robot in virtual environment.}
    \label{fig:robot_diag_sim}
\end{figure}

We consider a telepresence robot shown in Fig.~\ref{fig:robot_diag_sim} that is a differential drive robot comprising two driving wheels and four omnidirectional wheels for balance. 
The robot kinematics is given by the model
\begin{equation}
\label{eqn:unicycle}
\begin{alignedat}{2}
\dot{x}&=v\cos\theta \\
\dot{y}&=v\sin\theta \\
\dot{\theta}&=\omega,
\end{alignedat}
\end{equation}
in which $(x,y)$ is the robot position and $\theta$ is the orientation with respect to a global reference frame, and the control input $u=(v, \omega)$ corresponds to the linear and angular velocities with respect to the robot-fixed reference frame. The robot configuration is expressed as $q=(x,y,\theta)$ and it is constrained in the set $Q \subset \reals^2 \times S^1$. The control input is subject to actuation constraints (acceleration limits) and it takes part in a compact set of admissible controls $U \subset \reals^2$. 

Let $E \subset \reals^2$ be a planar environment in which the robot is moving. There are obstacles which are open sets and subsets of $E$ that prohibit the robot to have certain configurations due to collisions. The obstacles change dynamically and this information is (locally) available to the robot. % at discrete time instances $t_k$, $k=0, 1, \dots K$. % and $t_0$ is the initial time. 
Therefore, let $E_{obs}(t)\subset E$ be the union of all the obstacles known to the robot at time $t$. The part of the planar environment that the robot can be in without collisions at time $t$ is then denoted by $E_{free}(t)=E\setminus E_{obs}(t)$. Let $\robot: Q \rightarrow E$ be a mapping from the robot configuration to its footprint in the environment. The free configuration space is defined as $Q_{free}(t)=\{q \in Q \mid \robot(q) \in E_{free}(t)\}$. 
Let $q_g=(x_g,y_g,\theta_g)$ and $q_0=(x_0,y_0,\theta_0)$  be the goal and the initial configurations, respectively. Conventionally, the motion planning problem is defined as finding a control-trajectory $\tilde{u}:[0,T] \rightarrow U$ such that the state-trajectory $\tilde{q}: [0,T] \rightarrow Q$ computed as forward integrating \eqref{eqn:unicycle} starting from $q_0$ satisfies $\tilde{q}(t)\in Q_{free}(t), \forall t \in [0,T]$ and $\tilde{q}(T)=q_g$. Typically, a trajectory that is optimal with respect to a relevant metric is sought and the final time $T$ is not fixed.

As common in the literature, we consider that the motion planning of the robot is achieved by two interacting modules: the \emph{planner} and the \emph{controller}. The planner is responsible for computing a length-optimal path to the goal (not necessarily feasible with respect to the robot kinematics), and the task of the {controller} is to compute the control input that tracks the output of the planner. 
%that is optimal with respect to a relevant metric. 
We assume that a planner is in place and address the inclusion of the user input at the controller level. 
Before describing our problem, we will briefly explain the planner. Since the considered environment is dynamic, the planner is invoked at regular intervals $t_k$, $k=0, \dots, K$ to re-plan using the currently available information. Suppose the estimate of the robot configuration at $t_k$ is available and denote it by $\hat{q}_k=(\hat{x}_k, \hat{y}_k, \hat{\theta}_k)$. Then, given the current environment $E_{free}(t_k)$ and $\hat{q}_k$, the output of the planner is a length-optimal path $\pi_k : [0,1] \rightarrow E_{free}(t_k)$ such that $\pi_k(0)=(\hat{x}_k, \hat{y}_k)$ and $\pi_k(1)=(x_g,y_g)$. It is often referred to as the global path.
Note that the computed path is not kinematically feasible\footnote{A planner that computes kinematically feasible paths can also be used. In that case, $\pi$ maps to $Q_{free}(t_k)$ and satisfies \eqref{eqn:unicycle}}. However, it is assumed that the planner considers the robot size so that the path has sufficient clearance. Since the computational load of searching the whole position space is high, it is expected that the planner works at a lower frequency. Hence, it is mainly the task of the controller to avoid obstacles. 

%We address the problem of designing a controller that tracks the path computed by the planner taking the input given by the operator into account. To this end, 
We consider a controller inline with methods based on searching the control input space such as \ac{dwa} \cite{fox1997dynamic} and trajectory rollout \cite{gerkey2008planning}. These methods search a limited set of admissible controls (velocities) with respect to the actuation constraints such that given the current linear and angular velocities of the robot, only the ones that are achievable within a short length of time are considered. Consequently, assuming that the controls will stay constant, the respective state-trajectories are evaluated with respect to an objective function that captures the obstacle clearance and vicinity to the global path, for example. The control input corresponding to the best trajectory is then passed to the robot. This procedure is repeatedly executed for each $t$ considering $Q_{free}(t)$ and $\pi_k$ such that $t \in [t_k, t_{k+1})$.

Due to the computationally infeasible exhaustive search over the set of admissible controls, typically, only a set of sampled controls are passed to the trajectory evaluation step. Let $U^\Delta(t)$ be the set of sampled admissible controls at time $t$ and let $\Delta \tau$ be the time window used to integrate the dynamics. The set of control pairs such that respective trajectories are collision free are denoted as $U^\Delta_{free}(t)$. This implies that for each element $u \in U^\Delta_{free}(t)$ the trajectory $\tilde{q} : [t, t+\Delta \tau] \rightarrow Q$ computed considering a constant control for $\Delta \tau$ satisfies $\tilde{q}(\tau)\in Q_{free}(t), \forall \tau \in [t,t+\Delta\tau]$. 
Then, the best control input to pass to the robot at time $t$ is determined as 
\begin{equation}\label{eq:DWA_cost}
    u^*_t = (v_t^*, \omega^*_t) = \underset{{u_t \in U_{free}^\Delta(t)}}{\arg \min}\;  s_{nv}J_{nv}(u_t)
\end{equation}
in which $J_{nav}(u)=(J_1(u), J_2(u), \dots, J_d(u))^T$ is a $d$-dimensional vector of objective functions capturing the aspects relevant to path tracking, obstacle avoidance, and goal achievement and $s_{nav}=(s_1,s_2,\dots,s_d)$ is the respective vector of weights. %such that their multiplication yield a scalar cost. 

%The problem we are addressing is the ability to affect the robot's autonomous path. Our proposed method is a shared controller, where the user is able to adjust the path taken by an autonomous navigation but not taking full control so that the autonomous navigator will still take care of avoiding obstacles and finding the fastest route to the target location. 

%We address the problem of designing a controller that tracks the path computed by the planner taking into account the input given by the operator. 
We address the problem of designing a controller that not only tracks the path as described in the previous paragraph but also takes into account the input given by the operator. 
This corresponds to enabling the operator to apply slight modifications to the trajectory executed by the robot without directly controlling the robot motion. 
Suppose that the operator can indicate a preference for robot motion which is then mapped to a control command $u_h=(v_h,\omega_h)$. 
We propose a \ac{dwa}-based controller to incorporate the user input in the selection of the best control input pair to pass to the robot. To this end, we augment the minimization problem in \eqref{eq:DWA_cost} with an additional cost function that penalizes the difference between the selected control pair and the operator input. This way, the control input that is closer to the input given by the operator is preferred while taking into account other aspects related to navigation.
The cost function to evaluate the control input pairs and respective trajectories is described as 
\begin{equation}
    J(u_t) = 
    \begin{cases}
    s_{nv}J_{nv}(u_t) + s_{sh} J_{sh}(u_t)& \text{if } \gamma(t)=1\\
    s_{nv}J_{nv}(u_t) & \text{otherwise},
    \end{cases}
\end{equation}
in which $(v_h,\omega_h)$ is the pair of control inputs given by the operator at time $t$ and $\gamma$ is a function that maps $t$ to $1$ if user input is given and to $0$ otherwise. The cost component resulting from the user input is defined as $$J_{sh}(u_t) = \biggl(|v_{h} - v_t|, |\omega_h - \omega_t|\biggr)^T$$ and $s_{sh} = [s_v, s_\omega]$ is the vector of respective weights.
Consequently, the best control input in $U_{free}^\Delta(t)$ is determined as 
\begin{equation}\label{eq:DWA_human_cost}
    u^*_t = (v_t^*, \omega^*_t) = \underset{{u_t \in U_{free}^\Delta(t)}}{\arg \min}\; J(u_t).
\end{equation}

We introduce a delay in the transition to autonomous navigation to ensure that once the user input ceases, the resulting motion does not involve sudden rotations to compensate for the potential divergence from the global path due to user input. Let $t$ be the instance that the operator stops interacting with the robot. Then, a constant pseudo-input is given to the controller such that $v_h$ is the velocity command given at $t$ and $\omega_h=0$ for all $\tau \in [t, t+\delta]$, in which $\delta$ is the delay.

\section{Experiment} 

\subsection{Controllers}
The experiment is designed to compare three controllers. Two of them implement the shared control approach described in Section~\ref{sec:shared_control} with different input methods, and one implements a switch between the autonomous navigation and direct control of the operator.
Here we explicitly describe these methods. 

We integrate the controllers within the Nav2 project \cite{9341207} for \ac{ros2}. We use the default planner and localization plugins provided by Nav2 and modify the DWB controller which derives from \ac{dwa} to integrate ours. In particular, we use the default critics (objective functions forming the vector $J_{nv}$) and their default weights. Therefore,
\begin{equation*}
\begin{split}
    J_{nv}(u_t)= & s_{pa}\texttt{PathAlign}(u_t)+s_{pd}\texttt{PathDist}(u_t)\\
 &  + s_{bo} \texttt{BaseObstacle} +s_{ga}\texttt{GoalAlign}(u_t) \\
 & + s_{gd}\texttt{GoalDist}(u_t) + s_{rg}\texttt{RotateToGoal},
 \end{split}
\end{equation*}
in which for each trajectory corresponding to the numerical integration of the control $u_t$ for $\Delta\tau$, \texttt{PathAlign} ($s_{pa}=32.0$) and \texttt{PathDist} ($s_{pd}=32.0$) penalize the trajectory based on the distance from the global path and how well it is aligned to it, respectively. Similarly, \texttt{GoalAlign} ($s_{ga}=24.0$) scores a trajectory based on how well the robot aligns with the goal pose and \texttt{GoalDist} ($s_{gd}=24.0$) scores it based on how close the trajectory gets the robot to the goal pose.
\texttt{BaseObstacle} ($s_{bo}=0.02$) scores a trajectory based on its distance from the obstacles. Finally, it includes also a binary critic named \texttt{Oscillation} that prevents backwards-forwards motion by penalizing such trajectories with infinite cost. We refer the reader to Nav2 documentation~\cite{rosnav} for more detailed explanations.

% For the DWB local planner we use the default critics with their default scale values and our own shared control critic. The used critics with their scales and their job as described in the ROS Navigation2 documentation~\cite{rosnav}:
% \begin{itemize}
% \setlength\itemsep{0pt}
% \setlength\parskip{0pt}
% \item \texttt{PathAlign} ($s_{pa}=32.0$) gives the trajectory a score based on how well the robot aligns with the path.
% \item \texttt{PathDist} ($s_{pd}=32.0$) scores the trajectory based on how close the trajectory gets to the path.
% \item \texttt{BaseObstacle} ($Scale=0.02$) gives the trajectory a score based on the costmap it passes over.
% \item \texttt{GoalAlign} ($Scale=24.0$) gives the trajectory a score based on how well the robot aligns with the goal pose
% \item \texttt{GoalDist} ($Scale=24.0$) scores the trajectory based on how close the trajectory gets the robot to the goal pose.
% \item \texttt{RotateToGoal} ($Scale=32.0$) allows the robot to rotate to the goal orientation when it is sufficiently close to the goal location.
% \item \texttt{Oscillation} is binary critic that prevents the robot from just moving backwards and forwards
% \end{itemize}

%\textbf{Switching Control (SW)}
\subsubsection*{\textbf{\ac{sw}}} Switching control refers to handing the direct control of the robot to the operator when the operator wants to alter the course of the robot motion; similar systems have been proposed in \cite{chiou2016experimental} and implemented in the commercial Double 3 telepresence robot. For the remaining times, the robot is moving autonomously. The control input to send to the robot at time $t$, that is, $u^*_t$, is determined as 
\begin{equation}
    u^*_t = 
    \begin{cases}
    u_h=(v_h,\omega_h) & \text{if } \gamma(t)=1\\
    \underset{{u_t \in U_{free}^\Delta(t)}}{\arg \min}\;  s_{nv}J_{nv}(u_t) & \text{otherwise}.
    \end{cases}
\end{equation}
To make the switching explicit, we implemented a button such that once it is pressed the control is transferred to the operator. If after pressing the button no user input is given, the respective user input is $u_h=(v_h,\omega_h)=(0,0)$ and the robot stops. The user input is given using the joystick of an Oculus Quest 2 controller. Let $(p_x,p_y) \in [-1, 1]\times[-1, 1]$ be the joystick coordinate corresponding to the user input such that $(0,0)$ is the origin, it is mapped to $u_h=(v_h,\omega_h)$ as follows:
\begin{equation}
    v_h=
    \begin{cases}
        0 & \text{if } |p_y| \leq 0.1 \\
        p_y^2\sign(p_y)v_{max} & \text{otherwise},
    \end{cases}
\end{equation}
and
\begin{equation}\label{eq:map_to_wh}
    \omega_h=
    \begin{cases}
        0 & \text{if } |p_x| \leq 0.1 \\
        p_x^2\sign(p_x)\omega_{max} & \text{otherwise}.
    \end{cases}
\end{equation}

The following two methods implement the shared control described in Section~\ref{sec:shared_control}.
%\textbf{Shared Control-Joystick (SJ)}
\subsubsection*{\textbf{\ac{sj}}}
Similar to Switching Control, also in this case, the operator uses a joystick to provide input to the system. To keep controlling the robot simpler, we allow the user only to affect the rotational speed but not the linear speed. Therefore, for all $u_h$, $v_h$ is set to $v_{max}$ to avoid prioritizing controls corresponding to rotate in place and $w_h$ is computed using \eqref{eq:map_to_wh}. 
The control input $u^*_t$ passed to the robot at time $t$ is determined using \eqref{eq:DWA_human_cost}. 
To find a good combination of the relative weights, that is, $s_{sh}=(s_v, s_\omega)$ for penalizing the costs $J_{sh}(u_t)$ that take part in $J(u_t)$ and to determine a sufficient delay $\delta$,   
%To find out good values for the scales $s_v$, $s_w$ and smooth\_time for the shared control critic, 
we ran a pilot test ($N=8$). 
%From our previous experience we knew that the critic behaves well when $s_w=2s_v$. 
The participants tried this control method with four different parameter combinations, resulting in four conditions. For $(s_v,s_\omega,\delta)$ we tested the following four combinations: $(200, 400, 2s)$, $(400, 800, 2s)$, $(200, 400, 1s)$, $(400, 800, 1s)$.
%(see Table~\ref{tab:conditions} for parameters used in the pilot test). 
% \begin{table}%[h]
%     \centering
%     \begin{tabular}{l|l|l|l|}
%          \textbf{Condition}&  $s_v$ &  $s_w$ &  $\delta$\\\cline{1-4}
%          \textbf{1} & 200 & 400 & 2s  \\\cline{1-4}
%          \textbf{2} & 400 & 800 & 2s \\\cline{1-4}
%          \textbf{3} & 200 & 400 & 1s \\\cline{1-4}
%          \textbf{4} & 400 & 800 & 1s \\\cline{1-4}
%     \end{tabular}
%     \caption{The conditions of the pilot test.}
%     \label{tab:conditions}
% \end{table}
When asked which condition they felt was the best, 6 out of 8 participants picked condition 2, one participant picked conditions 3 and 4 as equally good, and 1 participant picked conditions 1 and 2 as equally good. Since condition 2 was selected by the majority of the users we used the parameters used in condition 2 for the main study ($s_{sh}=(s_v,s_\omega)=(400,800)$ and $\delta=2s$).

%\textbf{Shared Control-Gesture (SG)}
\subsubsection*{\textbf{\ac{sg}}} This method differs from \ac{sj} only in terms of the way the input is received from the operator in form of gesture control and a button to indicate the trajectory. When the operator presses the button, we obtain the current position and orientation of the hand from the tracking system of the Oculus Quest 2 controller. Then this is mapped to coordinates in the $x-y$ plane. The initial coordination of the hand is set as the origin of this plane and the normalized horizontal movement of the hand is mapped to $w_h$ similar to \eqref{eq:map_to_wh}.

% The projection of the hand position on the $x-y$ plane is set as the origin. While the button is held down, the  of the hand is mapped to $w_u$:

% \begin{equation}
%     w_h=
%     \begin{cases}
%         0,& \text{if Abs} (x_h)\leq 0.1\\
%         x_h*x_h*\text{Sign}(x_h)*w_{max}& \text{otherwise,}
%     \end{cases}
% \end{equation}
% in which $x_h$ is the hand's current horizontal position compared to the position where the button was originally pressed down.

\subsection{Hypotheses}
\label{sec:hypos}
We pre-registered the following two hypotheses, with the procedure and analyses to be used in the study, in Open Science Foundation (OSF) \url{https://osf.io/q9ubx}. In a scenario of a semi-autonomous telepresence robot navigating an environment in which users immersed in the robot through a head-mounted display (HMD) can alter the path executed by the robot either by switching to manual guidance or by indicating the direction that they would like to go, we will test the following hypotheses:
\begin{enumerate}[label=\textbf{H\arabic*:}]
    \itemsep=0em
    \item %H1:
    \ac{sj} condition is preferred as indicated by asking directly which condition was preferred.
    \item %H2:
    Perceived workload is lower under \ac{sj} as indicated by asking directly which condition was easier and NASA Task Load Index (NASA-TLX) questionnaire after each condition.
\end{enumerate}

\subsection{Study Setup}\label{sec:study_setup}
The hypotheses were tested using a simulated environment in Unity.
%We chose Unity because of its extensive support for virtual reality devices. 
The virtual environment was loosely based on the University of Oulu campus, so although some participants could recognize parts of the environment, they could not use this knowledge into their advantage. 
The simulated telepresence robot used in the experiment comprised of a mobile base as described in Section~\ref{sec:shared_control} and a simulated 360$^\circ$ camera attached 1.5 meters above the robot from which the users could see the virtual world using a virtual reality headset.
We also ensured that the dynamics governing the robot motion in simulation were sufficiently realistic. 
The autonomous navigation of the robot was based on the Robot Operating System (ROS) and its Nav2 project \cite{9341207}. To avoid high computational loads that might hamper the viewing comfort, we used two separate computers; a ROS-based application that ran on a Linux laptop and Unity-based one on a Windows laptop. The connection was established via a ROS TCP Connector\cite{UnityRos} (Fig.~\ref{fig:teleop}).
In the experiment, the robot was navigating towards a fixed goal configuration using the autonomous navigation system together with the controllers described in the previous section. 
The virtual environment contained regions called \emph{regions to avoid} that consisted of potholes, scaffolding, cardboard boxes, traffic cones, and bumpy areas. These were not visible to the robot's sensors and were not marked on the map so that they could not be avoided like other obstacles. However, they were passable for the robot. Fig.~\ref{fig:robot_diag_sim}b shows an instance from the experiment with traffic cones and potholes. 

%There are obstacles like potholes, scaffolding, cardboard boxes and traffic cones that the robot can't detect and the subject has to alter the path the robot would take to get past these obstacles. There is also some bumpy tiles on the floor which doesn't prevent the robot from moving on them but they make the user uncomfortable. The subjects are given a task to avoid these obstacles and help the robot to reach the goal as fast as possible

%The simulated mobile robot consists of a platform with two differential drive motors in the middle and four caster wheels, one in each corner. The participants can see around using a Virtual Reality headset showing video feed from simulated 360-degree camera attached 1.5 meters above the robot.

\subsection{Procedure}
The participants were presented with three conditions corresponding to the \ac{sw}, \ac{sj}, and \ac{sg} control methods. The conditions \ac{sw} and \ac{sj} were presented in counterbalanced order such that both videos were seen first and second an equal number of times by the participants, and the \ac{sg} condition was presented always as the last one. 
Upon arrival, the participants were greeted by a researcher and signed a form to indicate their consent to participate. Next, the experimenter asked the participants if they were feeling nauseous or had a headache in an effort to pre-screen people feeling sick already before the experiment began. 
Then, the participants were told the general instructions by the experimenter, and shown how to put on the HMD. 

After the experimenter made sure that the participant knew how to put on the HMD properly, they read out the instructions for the first task. The participants were told that they were late for a meeting where they were going to participate using a telepresence robot. The robot is capable of navigating autonomously, that is, computing a path to a goal and tracking while avoiding obstacles.
The goal was set as a point in the meeting room and to have more control over the experiment, the users were not allowed to change the goal of the robot. There were potholes and scaffolding along the robot's default path without user deformations, which are not visible to the robot’s sensors. 
Despite not being explicitly told to avoid those regions, the participants were encouraged to do so by saying that they could get slowed down or feel uncomfortable (bumpy tiles) if they did not. They were also told that the path the robot would try to follow was shown as a green line on the floor.
Before each task, the participant was asked to practice the controls in a short practice environment, with the robot again following a path with similar scaffolding obstacles as in the real path, lasting approximately 2-3 minutes.

After the practice, the participants were asked if they are ready for the task, and the proper task scenario was started by the experimenter. After the task, the participants were asked to take off the HMD and fill out a questionnaire regarding their experience with that specific control method. The same procedure was repeated two more times with the other control methods. 
Finally, the participants were given 20€ Amazon vouchers for participating. After each experiment, all the equipment was disinfected as a precaution regarding COVID-19. Precautions were also taken during the experiment by using disposable face covers with the HMD, the experimenter always wearing a mask, and the experimenter keeping a safe distance from the participant except if help was needed.

%you need to get to a meeting on the other side of the building. You are late and still at home, so you decided to use the telepresence robot to get there faster. The robot can find a path there on it’s own (it will move autonomously, like a self-driving car) but there are renovations going on. There are potholes and scaffolding in the way of the path which are not visible to the robot’s sensors. You can see the path the robot will try to follow as a green line on the ground. You need to help the robot to avoid the obstacles using your controller so you aren’t slowed down by them. You can also alter the robot’s path when preferred, for example if you prefer going further away from people or corners than the autonomous path would take you. There are also some uneven floor materials, like dark blue tiling, that will be uncomfortable to ride on so you might want to steer away from those.
%The participants were told that they were late to get to a meeting in order to compare how fast they got to the goal using each method. 

\subsection{Measures}
For each task, we measured the path that the robot took. To see where the subjects altered the path of the robot, we placed time stamps on the moments when the subject used their controller. We also measured the head movements of the subjects using the Oculus Quest 2's tracking system to see where the subjects were looking at any point during the task.  

At the beginning of each questionnaire, we asked the participants to fill out a Simulator Sickness Questionnaire (SSQ) \cite{ssq}. It is a questionnaire often used to measure the sickness that results from using VR, and consists of questions regarding 16 different sickness symptoms that are scored based on severity of experience. Weighted scores are used to calculate the total score for the sickness. Higher scores indicate greater levels of sickness experienced. Participants were also asked to fill out a NASA Task Load Index questionnaire, which is used to measure six dimensions of workload (mental demand, physical demand, temporal demand, performance, effort, and frustration) of the task. Each dimension is compared between the methods individually to find if one of the methods is perceived as more demanding than the others. These questions were followed by a forced-choice question if there was any point where the participant wanted to alter the path but didn't, and 7-point Likert-scale questions about the participant's perceived control over the robot, the ease of altering the path, and their comfort while altering the path.

After the second task, we additionally asked forced-choice questions about the participants' preference between the two methods, and a comparison between the two methods on their control and ease of use. We also asked open-ended questions about the reasons for their choices and if there were any specific situations where they would prefer one of the methods over another.

Finally, after the third task, we asked forced-choice questions about whether the \ac{sg} method made them feel more as if they were there in the environment or if they felt more in control of the robot than in the previous tasks. We also asked their preference over all three different methods and open-ended questions about reasons for their choices. In the end we asked about participant's VR and gaming experience and their demographics. The questionnaire after the third task, including also all previously asked questionnaires except the forced-choice between first two methods, can be found from the same OSF page as the hypotheses (Section~\ref{sec:hypos}).

\subsection{Participants}
Participants were recruited from the University of Oulu campus and community. We aimed to have 32 participants, but due to the exclusion of four people from the study, we ended up running 36 total. Three of the excluded participants quit the experiment due to sickness symptoms, and one was excluded for not following the instructions and moving the robot outside of the intended environment, which caused them to get stuck and thus they were not able to finish their task without a reset. Of the 32 included participants, 19 were men and 13 were women. %All participants reported having normal or corrected-to-normal vision and none of the participants were colorblind. 
The responses of the participants to how often they use \ac{vr} systems were: $21.9\%$ never used any before, $43.8\%$ just a couple of times and at least once, $18.8\%$ once or twice a year, $9.4\%$ once or twice a month, and $3.4\%$ once or twice a week.

\section{Results}
Two confirmatory hypothesis tests were preregistered. All tests were run in SPSS with significance levels set to $0.05$ and with a $95 \%$ confidence interval.

\subsection{Confirmatory analysis}

\begin{figure}%[t!]
\centering
\includegraphics[trim=0 19 0 15,clip,width=0.7\linewidth]{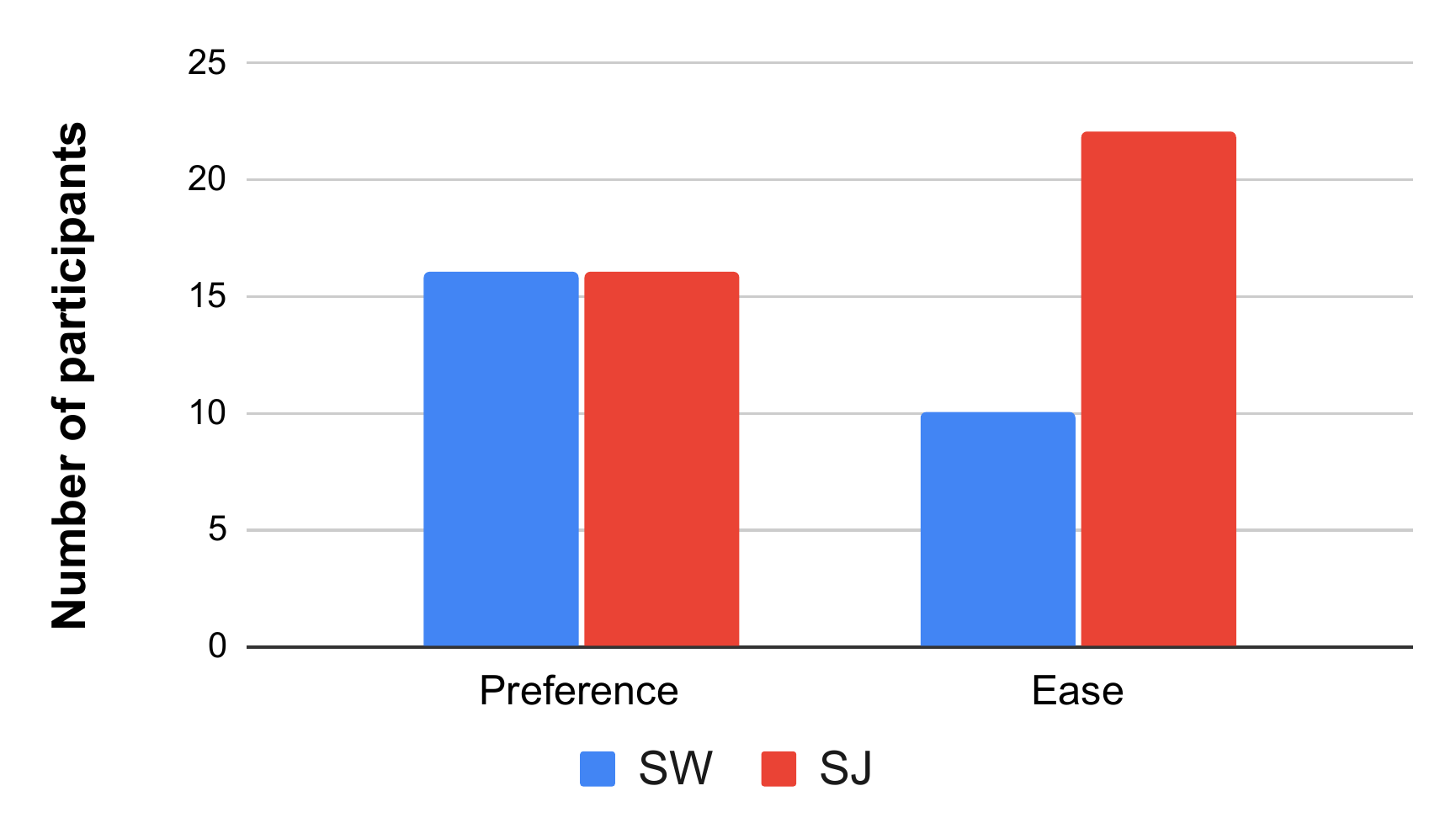}
\caption{The distributions of responses to the questions regarding preference, easiness, and comfort.}
\label{fig:preference_and_ease}
\vspace{-0.6cm}
\end{figure}

Fig.~\ref{fig:preference_and_ease} shows the distributions of the responses given by the participants to the forced-choice questions regarding preference and ease of use, and comfort. 
When asked ``Which control method did you prefer?" 16 out of 32 participants $(50\%)$ selected the \ac{sj} condition showing no tendency in either direction in preference.  
%An exact binomial test with exact Clopper-Pearson $95\%$ CI indicated that this tendency in preference is not statistically significant, $p=1.000$ (two-sided) and had a $95\%$ CI of $31.9\%$ to $68.1\%$.  
When asked ``Which control method was easier to use?" 22 out of 32 participants $(68.75\%)$ selected the \ac{sj} condition. An exact binomial test with exact Clopper-Pearson $95\%$ CI was performed, showing that the condition \ac{sj} was found significantly easier compared to the \ac{sw} condition, $p=0.026$ (one-sided) and had a $95\%$ CI of $50.0\%$ to $83.9\%$
%the shared joystick method was (borderline?) significantly easier than the switching method, $p=0.052$ (two-sided) and had a $95\%$ CI of $50.0\%$ to $83.9\%$.

\begin{figure}
\centering
\subfigure[]{\includegraphics[trim=75 445 75 60,clip,width=0.49\linewidth]{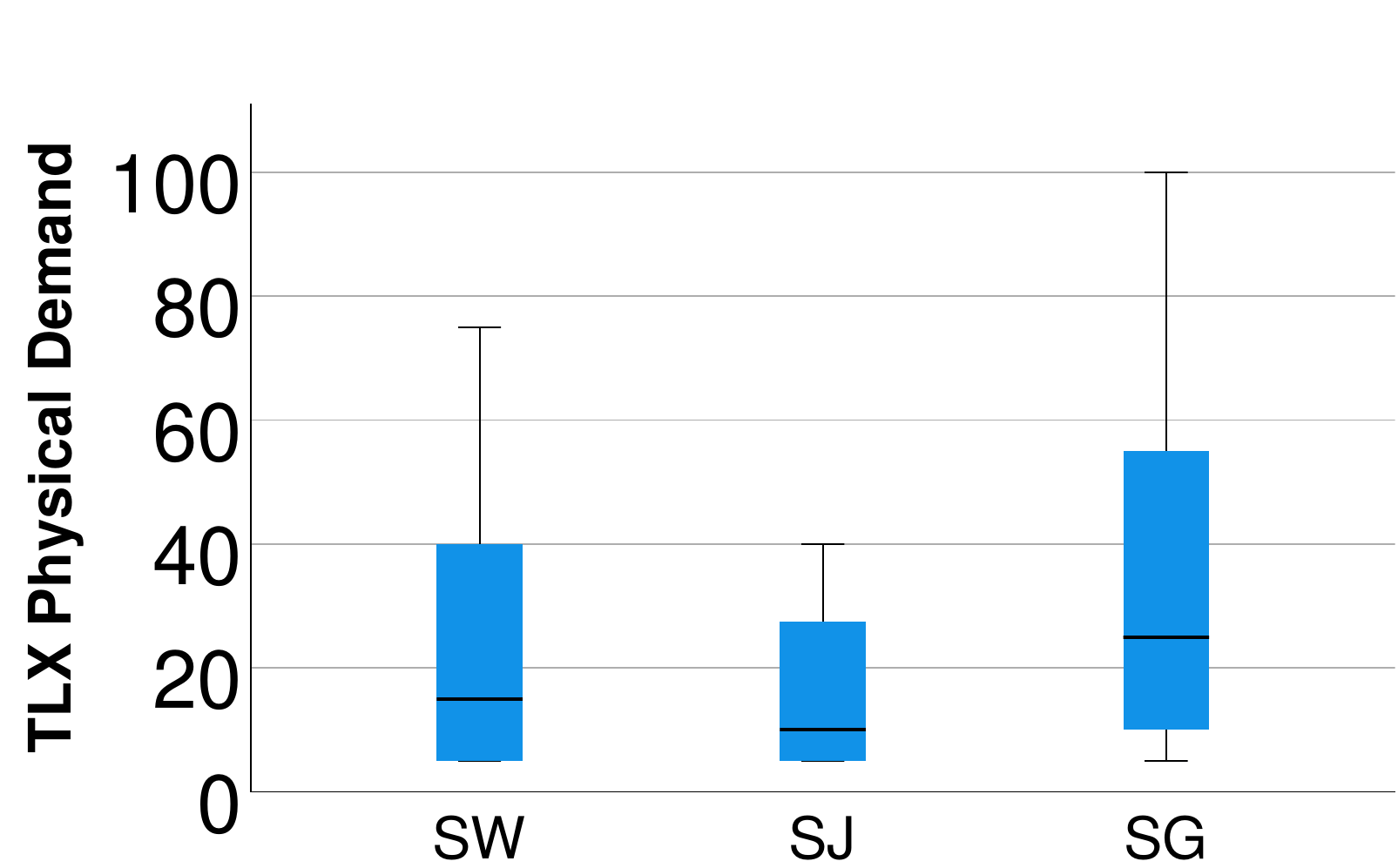}}
\subfigure[]{\includegraphics[trim=75 445 75 60,clip,width=0.49\linewidth]{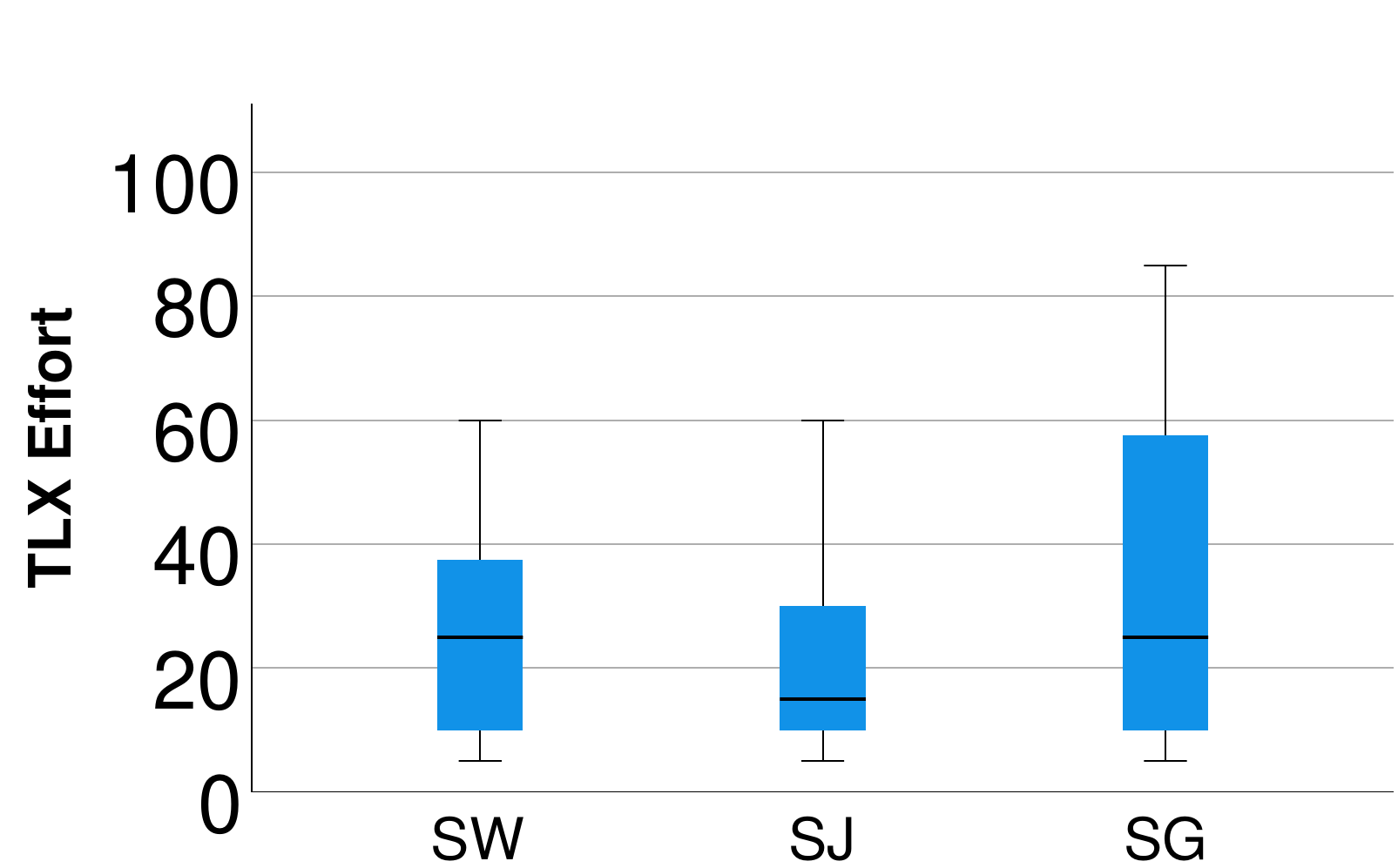}}
\caption{Comparison of relevant TLX scores (a) physical demand (b) effort.}
\label{fig:TLX}
     \vspace{-0.55cm}
\end{figure}

We compared the differences between the TLX-scores across conditions for each dimension of workload and found a statistically significant difference only in the effort dimension. A Wilcoxon Signed-Ranks test is performed to compare the TLX effort scores for \ac{sj} (${Mdn}=15.0$) and \ac{sw} (${Mdn}=25.0$) (see Fig.~\ref{fig:TLX}b for the score distributions).
The test indicated that 
%there is weak evidence that 
\ac{sj} elicited statistically significantly lower effort scores compared to \ac{sw}, $Z=-1.687$, $p=.047$, $r=0.30$ (one-sided).

%Consequently, when a Wilcoxon Signed-Ranks test (one-sided) is performed to compare the differences between the TLX effort scores for \ac{sj} (${Mdn}=15.0$) and \ac{sw} (${Mdn}=25.0$) conditions (see Fig.~\ref{fig:TLX_eff} for the score distributions). The test indicated that there is evidence that the \ac{sj} elicited lower effort scores compared to \ac{sw}, $Z=-1.687$, $p=.047$, $r=0.30$ (one-sided). Other TLX scores didn't have any significant differences between these methods.

\subsection{Exploratory analysis}
In addition to the confirmatory analyses, we performed exploratory analyses to interpret the results better.

\subsubsection{Quantitative}

The task completion time data was analyzed to see if one method would result in faster task completion. The respective distributions for \ac{sj} and \ac{sw} conditions did not follow a normal distribution as indicated by a Shapiro-Wilk test, $W=0.773$, $p<0.001$ and $W=0.852$, $p<0.001$, respectively. Therefore, a Wilcoxon Signed-Ranks test (two-sided) was performed to compare the differences between the task completion times for \ac{sj} ($Mean=172.37s$) and \ac{sw} ($Mean=187.17s$) conditions. The test indicated that \ac{sj} elicited significantly faster task completion compared to \ac{sw}, $Z=-4.207$, $p=<0.001$, $r=0.74$.

%The distributions were significantly non-normal for the task completion times of the \ac{sj} method ($W=0.773$, $p<0.001$), the \ac{sw} method ($W=0.852$, $p<0.001$), and the \ac{sg} method ($W=0.684$, $p<0.001$) according to Shapiro-Wilk tests. 
%Comparing the time taken to finish the task using \ac{sj} method ($Mean=172.37s$) with the time taken to finish the task using \ac{sw} method ($Mean=187.17s$), we found that the \ac{sj} method had significantly lower finishing times, as indicated by a Wilcoxon Signed-Ranks test (two-sided), $Z=-4.207$, $p=<0.001$, $r=0.744$.

To see if one control method helped to avoid \emph{regions to avoid} more, we analyzed the number of regions that the path executed by the robot intersected with. Comparing the number of regions not avoided for \ac{sj} ($Mean=0.219$) with \ac{sw} ($Mean=0.406$) we did not observe any significant difference as indicated by a Wilcoxon Signed-Ranks test (two-sided), $Z=-1.281$, $p=0.213$, $r=0.226$. Considering the obstacles, in total, we observed three collisions under \ac{sw} condition. There were no collisions for \ac{sj} as it is inherently collision-free.

%To see if one control method helped to avoid obstacles better, we analyzed the number of collisions into obstacles not visible to the robot's sensors. When comparing the number of these collisions when using the \ac{sj} method ($Mean=0.219$) with the number of collisions when using the \ac{sw} method ($Mean=0.406$), there was no significant difference, as indicated by a Wilcoxon Signed-Ranks test (two-sided), $Z=-1.281$, $p=0.213$, $r=0.226$. When comparing the collisions with the walls that were visible to the robot's sensors, subjects using the \ac{sj} method had no collisions at all while subjects using the \ac{sw} method had 3 among all the participants. However, this difference was not significant when compared with a Wilcoxon Signed-Ranks test (two-sided), $Z=-1.732$,

We tested whether one condition induced more sickness. Comparing the weighted total SSQ scores across the conditions \ac{sj} ($Mean=44.5294$) and \ac{sw} ($Mean=48.3863)$, we did not observe a significant difference between the scores, as indicated by a Wilcoxon Signed-Ranks test (two-sided), $Z=-0.809$, $p=0.427$, $r=0.14$. However, we observed a carryover effect on sickness with 15 minutes breaks; total weighted SSQ scores after the task 2 ($Mean=54.1131$) was significantly higher compared to the task 1 ($Mean=38.80$), as indicated by a Wilcoxon Signed-Ranks test (two-sided), $Z=-3.648$, $p=<.001$, $r=0.64$.

To find out the reasons for participants' choices for their preference between the \ac{sw} and \ac{sj} methods, we filtered the participants into two groups based on their preference. 
We then compared the Likert-scale ratings for the easiness and the feeling of control over the robot within those groups. 
From the participants who preferred the \ac{sw} method, when comparing their ratings of the control over the robot, we found out that there is, at best, weak evidence that the \ac{sw} method ($Mean=1.9375$) had more control over the robot than the \ac{sj} method ($Mean=2.8750$), as seen from a Wilcoxon Signed-Ranks test (two-sided), $Z=-1.943$, $p=0.052$, $r=0.343$. However, when comparing the easiness ratings within these participants, there was no significant difference between the \ac{sw} method ($Mean=2,0625$) and the \ac{sj} method ($Mean=2.5625$) as seen from a Wilcoxon Signed-Ranks test (two-sided), $Z=-1.144$, $p=0.253$, $r=0.202$. From the people who preferred the \ac{sj} method, a Wilcoxon Signed-Ranks test indicated that there was no significant difference between the means of the control over the robot ratings between the \ac{sw} method ($Mean=2.3750$) and the \ac{sj} method ($Mean=2.4375$),  $Z=-0.265$, $p=0.791$, $r=0.047$, but there was again, at best, weak evidence that the easiness score was lower with the \ac{sj} method ($Mean=1.6250$) than with the \ac{sw} method ($Mean=2.3125$), $Z=-1.942$, $p=0.052$, $r=0.343$.Similarly, %as seen on Table~\ref{tab:pref-ease}, 
of the participants who preferred the \ac{sw} method, only 10 out of 16 ($62.5\%$) said that it also felt easier, while all 16 participants who preferred the \ac{sj} method also picked it as easier method.

%\begin{table}[h]
%    \centering
%    \begin{tabular}{l|l|l|}
%        &  \textbf{Percieved \ac{sw} easier} &  \textbf{Percieved \ac{sj} easier}\\\cline{1-3}
%        \textbf{Preferred \ac{sw}} & 10 & 6 \\\cline{1-3}
%        \textbf{Preferred \ac{sj}} & 0 & 16 \\\cline{1-3}
 %   \end{tabular}
%    \caption{...}
%    \label{tab:pref-ease}
%\end{table}

With our exploratory third control method, we wanted to see if participants would prefer using their gestures (hand) to give inputs to the shared controller instead of the joystick. 
%The \ac{sg} method seemed more demanding, . 
When we asked "Which of the three methods did you prefer?", the \ac{sg} method was preferred by 15 out of 32 participants ($46.9\%$) while the \ac{sw} method was preferred by only 8 ($25.0\%$) and the \ac{sj} method was preferred by only 9 participants ($28,1\%$). 
When asked "How easy was it to alter the path of the robot", participants felt that the \ac{sj} method was significantly easier to use ($Mean=2.0937$) than the \ac{sg} method ($Mean=2.9687$), as indicated by a Wilcoxon Signed-Ranks test (two-sided), $Z=-2.381$, $p=0.015$, $r=0.421$. When comparing the physical demand score between the \ac{sj} method ($Mean=3.969$) and the \ac{sg} method ($Mean=7.031$), the \ac{sg} method had significantly higher score, as indicated by a Wilcoxon Signed-Ranks test (two-sided), $Z=-2.785$, $p=0.004$, $r=0.492$. Similarly the participants felt that the \ac{sg} method required more effort ($Mean=6.969$) than the \ac{sj} method ($Mean=4.875$), as indicated by a Wilcoxon Signed-Ranks test (two-sided), $Z=-2.034$, $p=0.041$, $r=0.360$. Other TLX scores did not have significant differences. TLX score distributions can be seen in Fig.~\ref{fig:TLX}.
%On top of forced-choice question about the control over the robot, Likert-scale is used to measure the control over, easiness and the comfort. Comparing the control over the robot in \ac{sg} condition ($Mean=2.6563$) with the ones in \ac{sj} condition ($Mean=2.66$), we did not find significant difference, as indicated by a Wilcoxon Signed-Ranks test (two-sided), $Z=-1.450$, $p=0.153$, $r=0.256$. When asked ``Did you feel more in control of the robot than in the previous sessions?" the answers were divided equally 16 participants saying yes and 16 participants saying no. An exact binomial test with exact Clopper-Pearson $95\%$ CI was performed, showing no significance, $p=1.000$ (two-sided) and had a $95\%$ CI of $31.9\%$ to $68.1\%$. Similarly there was no significant difference between the comfort scores between the \ac{sh} condition ($Mean=4.3438$) and the \ac{sj} condition ($Mean=4.3750$), as indicated by a Wilcoxon Signed-Ranks test (two-sided), $Z=-0.115$, $p=0.915$, $r=0.02$. However, when comparing the easiness scores of the \ac{sh} condition ($Mean=2.9687$) with the \ac{sj} condition ($Mean=2.0937$), we found that the \ac{sj} method felt significantly easier to the participants than the \ac{sh} method, as indicated by a Wilcoxon Signed-Ranks test (two-sided), $Z=-2.381$, $p=0.015$, $r=0.42$. Interestingly when we asked the preference out of all three methods the \ac{sh} method was preferred by 15 out of 32 participants ($46.9\%$) while the \ac{sw} method was preferred by only 8 ($25.0\%$) and the \ac{sj} method was preferred by only 9 participants ($28,1\%$).

\subsubsection{Qualitative}

\begin{figure}
    \centering
    \subfigure[]{\includegraphics[trim=2 3 5 2,clip,width=0.49\linewidth]{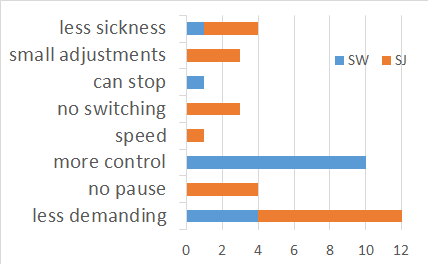}}
    \subfigure[]{\includegraphics[trim=2 28 5 2,clip,width=0.49\linewidth]{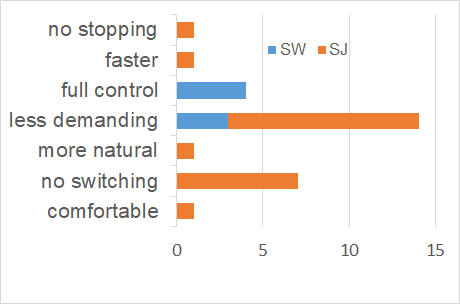}}
    \caption{Frequently found codes for questions (a) ``Please explain why you prefer that control method”. (b) ``Please explain why that control method felt easiest to use”}
    \label{fig:pref_oe_responses}
    \vspace{-0.35cm}
\end{figure}

% \begin{figure}
%     \centering
%     \includegraphics[trim=2 3 2 2,clip,width=0.8\columnwidth]{figures/pref_of_3_bins.png}
%     \caption{Frequently found codes for question ``Please explain why you prefer that control method.”.}
%     \label{fig:pref_3_oe_responses}
%     \vspace{-0.55cm}
% \end{figure}

% \begin{figure}
%     \centering
%     \includegraphics[trim=2 3 2 2,clip,width=0.6\columnwidth]{figures/ease_of_use_bins.png}
%     \caption{Frequently found codes for question ``Please explain why that control method felt easiest to use”.}
%     \label{fig:ease_oe_responses}
%     \vspace{-0.55cm}
% \end{figure}

The open-ended data was analyzed using the thematic analysis method with inductive approach \cite{patton2005qualitative}. %In the first stage of analysis, two researchers independently identified codes from the response data and mutually agreed on the codings in the second phase. It was not mandatory for participants to answer the open-ended questions, and these fields were often left blank.

Fig.~\ref{fig:pref_oe_responses}a shows the frequently found codes in the responses to the open-ended question asking why participants preferred the control method, divided by which method was preferred. The most frequently found code was \emph{less demanding}. Eight ($50\%$) participants who preferred \ac{sj} and four ($25\%$) participants who preferred \ac{sw} found these methods less demanding (``\textit{It felt that it would be less laborious to just make small adjustments to the path when needed rather than drive manually}"). The biggest factor for participants who preferred \ac{sw} was having \emph{more control} that is found in the responses of $10$ out of $16$ participants (``\emph{With the first method it was always clear who was in control. The second method felt like a constant struggle.}"). 

The responses to the open-ended questions regarding why the control method felt easiest to use the greatest number of comments ($44\%$) said that it was \emph{less demanding}, followed by \emph{no switching} ($22\%$) and \emph{full control} ($13\%)$. See Fig.~\ref{fig:pref_oe_responses}b for the frequently found keywords in the participants' responses.

\begin{figure}
    \centering
\subfigure[]{\includegraphics[trim=0 0 0 0,clip,width=0.49\linewidth]{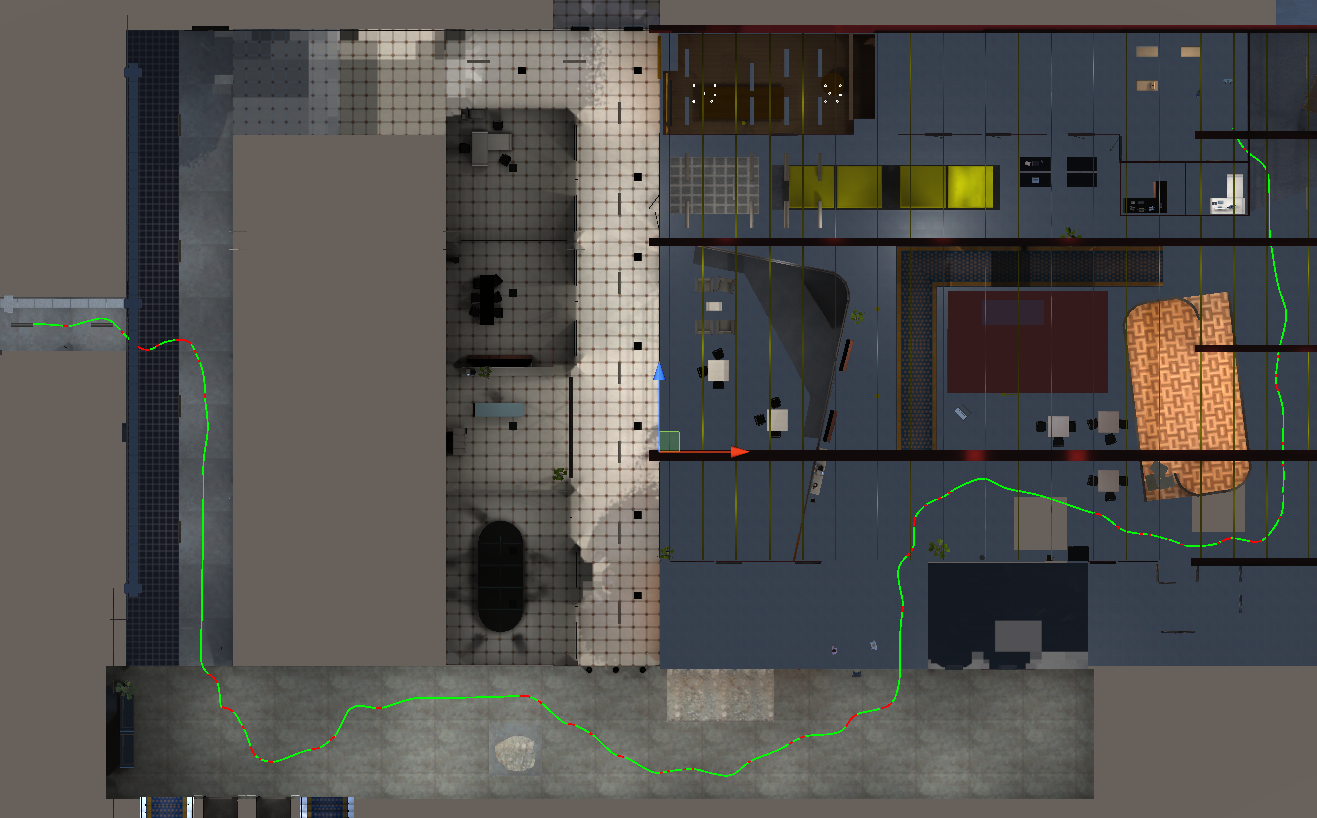}}
\subfigure[]{\includegraphics[trim=0 0 0 0,clip,width=0.49\linewidth]{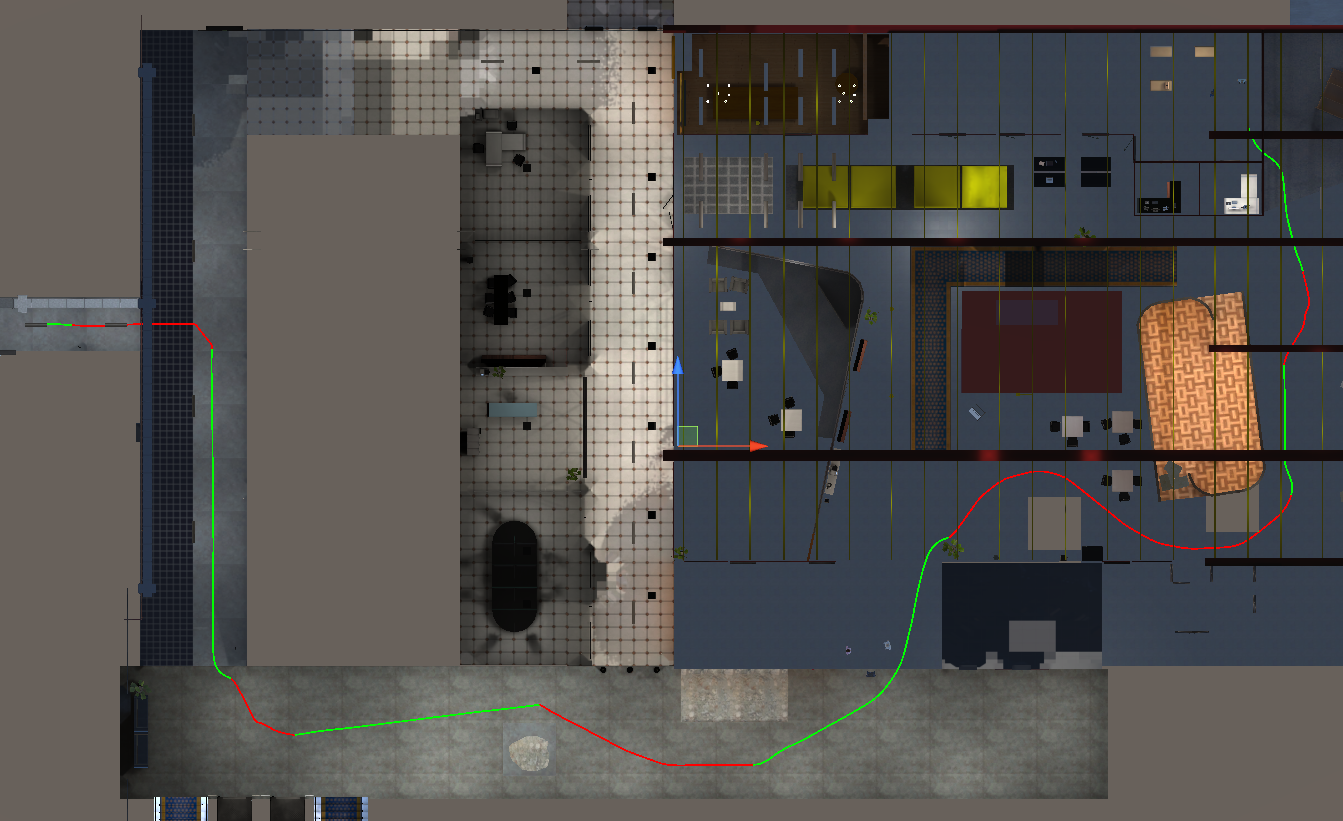}}
\caption{Example paths taken using (a) \ac{sj} and (b) \ac{sw}. Red parts are where the participant used their controller to give input.}
\label{fig:sj_sw_paths}
\vspace{-0.4cm}
\end{figure}

\section{Discussion}
An interesting takeaway from the performed study was the discrepancy between ease of use and preference: even though preference between \ac{sj} and \ac{sw} was split exactly equal, participants found \ac{sj} statistically significantly easier compared to \ac{sw}. Whereas it was expected that experienced gamers not to have any issues with manual joystick control over such a short time period (approximately 3 minutes), we did not observe any correlation between the gaming background and preference of control method
%gaming background did in fact not have a correlation with the preference between methods
, even though several participants explicitly said that it did (\textit{``It did not demand much mind effort to accomplish. and having joystick in the hand is a good experience from past playing video games."}). 

One reason why participants preferred \ac{sw} but found \ac{sj} easier was the \textit{feeling of control}. For example, one participant preferred \ac{sw} because \textit{``You can have full control and navigate the path from the location of your choosing"}, but found shared control easier because \textit{``its easier because you can relax and see where you want to change path. However on like the first, you can feel sleepy or over relaxed with the second"} for whom the second condition was \ac{sj}.
%(second" was shared control). 
This result could be related to our study setup, 
for which the goal configuration was constant throughout the experiment;
%where the goal of the robot's autonomous path stayed the same throughout the whole experiment; 
in a real scenario, users would have chosen the goal themselves, either via waypoint navigation (see Section~\ref{sec:intro}) or via choosing the destination on a minimap. 

Another frequently mentioned notion, related to the feeling of control, was the ability to stop. Whereas taking full manual control of the robot stopped the robot, with shared control, the robot always drove towards the goal. Besides not including waypoint navigation, this was another deliberate choice for the study setup: we were worried about too much freedom of choice for the participants to confound the study with multiple simultaneous locomotion modalities. However, if they had had more control over the robot's %autonomous 
destination and stopping, there is a chance users would have felt more in control even in the \ac{sj} condition.
%setting. 
Also, with waypoint navigation, the participants could have chosen whether to use shared control or waypoint navigation depending on the size of the obstacle: however, giving them this freedom would have made analysis of the results more difficult. Several participants noticed that the proposed method was especially useful for small corrections \textit{``It felt that it would be less laborious to just make small adjustments to the path when needed rather than drive manually.}", which is indeed the intended use case. Fig.~\ref{fig:sj_sw_paths} presents an example of this case such that \ac{sj} is used to make small alterations to the path whereas \ac{sw} is used along larger portions of the path executed by the robot.

%Shared hand had similar completion time as shared joystick, but due to counterbalancing we don't report the results.

%SSQ scores were really quite high when compared to previous runs. 

Since the main contribution of this paper is the underlying shared control mechanism, which can be used with either immersive or regular telepresence robots, the gesture condition, mainly meant for immersive telepresence robots, was left last as exploratory. However, the results were encouraging; whereas using extensive gesture motions for control is not encouraged in VR, due to fatigue, in this case the participants had armrests to rest their arms on, and the control is planned to be needed only on occasions. Also, the joystick of the \ac{hmd} controller is quite small and not very accurate, which could have been difficult to use for non-gamers. There are also interesting reasons in the qualitative data for preferring the gestures, such as \textit{``It was more comfortable, a lot easier and it felt more real"}. The "felt more real" part could be related to \textit{presence}, which is one of the main reasons for using \ac{hmd} for telepresence. Thus, future research on whether body-based control increases the feeling of presence would be interesting. 

%Funny things: participants WERE faster in the shared control, and also reported that the robot felt faster. But that's likely due to not being able to stop, and control etc. 
%Could we use this with others besides DWA?

\section{Conclusion}
In this paper we presented \ac{hidwa}, a novel shared control method based on DWA aimed especially for telepresence robots such that the users can influence the control input to the robot resulting in trajectories according to their choice. 
%In this paper we presented \ac{hidwa}, a novel shared control method based on DWA aimed especially for telepresence robots, where users can influence the autonomous motion planner to choose trajectories in a direction of their choice. 
We performed a user study in VR, where the users avoided regions unseen by the robot's sensors either with the proposed method or switching between autonomous and manual modes. We showed that the participants found shared control easier, even though preference was split equally between the proposed method and switching; the feeling of control was often mentioned as the reason for users who preferred the switching, even though there was no statistically significant difference in a Likert-scale question of feeling of control. In the future study, we will allow the users to select the goal. We expect it to increase the feeling of having control over the robot and thus, affect their preference for different control methods.
%This may increase the feeling of control for the users and thus also affect preference on top of ease.

%also affect the final goal of the robot, whereas in this study the autonomous path was predetermined and only partial changes were possible; this may increase the feeling of control of the users and thus also affect preference on top of ease.
%We performed a user study in VR, where the users avoided regions unseen by the robot's sensors either with the proposed method or with the ability to switch between autonomous motions and fully manual joystick control. We showed that the participants found shared control easier, even though preference was split equally between the proposed method and switching; the feeling of control was often mentioned as the reason for users who preferred the switching, even though there was no statistically significant difference in a Likert-scale question of feeling of control. In future work we want to try our system in a setting where the users can also affect the final goal of the robot, whereas in this study the autonomous path was predetermined and only partial changes were possible; this may increase the feeling of control of the users and thus also affect preference on top of ease.

\bibliographystyle{IEEEtran}
\bibliography{refs}

\end{document}